\documentclass{article}

     \PassOptionsToPackage{numbers, compress, square, comma}{natbib}

\usepackage[preprint]{neurips_2020}

\usepackage{float}
\usepackage{amsmath} 
\usepackage[utf8]{inputenc} 
\usepackage[T1]{fontenc}    
\usepackage{hyperref}       
\usepackage{url}            
\usepackage{booktabs}       
\usepackage{amsfonts}       
\usepackage{nicefrac}       
\usepackage{pifont}
\usepackage{enumitem}
\usepackage{adjustbox}
\usepackage{siunitx}

\usepackage{subcaption}
\usepackage{caption}
\usepackage{ragged2e}
\usepackage{array,multirow}

\usepackage[table]{xcolor}

\definecolor{color-A65B73}{HTML}{A65B73}
\definecolor{color-EB9706}{HTML}{EB9706}
\definecolor{color-4A9995}{HTML}{4A9995}
\definecolor{color-204E61}{HTML}{204E61}

\usepackage{graphicx}
\newcommand{\nfmap}{\textsc{NFMap}}

\makeatletter
\def\fixedlabel#1#2{%
  \@bsphack%
  \protected@write\@auxout{}%
         {\string\newlabel{#1}{{#2}{\thepage}}}%
  \@esphack}
\makeatother

\title{NFMap: Topographic 2D Manifold Learning with Invertible Normalizing Flows}

\author{
  H\'el\'ena Alexandra Gaspar
  \\
  LovingScience\\
  124 City Road, London EC1V 2NX\\
  \texttt{helena@lovingscience.com}
}

\begin{document}

\maketitle
\setcounter{footnote}{0}

\begin{abstract}
We present \nfmap{}, a dimensionality reduction method for 2D visualisation that
extends Generative Topographic Mapping (GTM) with a bijective normalizing flow.
Like GTM, \nfmap{} places a fixed mixture grid prior over a 2D latent map, giving a parametric embedding, an explicit likelihood, and the ability to project new points onto an existing map; the likelihood is an effective out-of-distribution signal on chemical datasets. Unlike GTM, whose latent-to-data mapping is an RBF network of bounded flexibility, \nfmap{}'s mapping is a normalizing flow, and is therefore fully invertible: any location on the map can be carried back to data space. To alleviate map crowding on large datasets, \nfmap{} additionally supports heavy-tailed Student-t mixture components.

\nfmap{} maps data to a latent space of the same dimension, split into a 2-dimensional navigable map and an off-manifold residual. Because the flow is bijective and its objective penalises volume contraction, variance squeezed out of the residual must reappear in the navigable map. The residual scale is the main control between local neighbourhood fidelity and global distance fidelity; its effect depends on the dimensionality of the data and is coupled to the extent of the latent grid.

\end{abstract}

\section{Introduction}
\vspace{-0.3cm}

Neighbour-embedding methods such as t-SNE~\citep{tsne} and UMAP~\citep{umap} preserve local structure well and are the default in most pipelines. However, they are non-invertible. Their standard versions are non-parametric, but both have parametric counterparts~\citep{parametric_tsne, parametric_umap}.

Generative topographic mapping~\citep{gtm} (GTM) fits a constrained Gaussian mixture whose centres lie on a fixed 2-D grid, giving a parametric, probabilistic, out-of-sample-capable map. It is the direct ancestor of this work and remains strong on global structure. Its limit is representational: the latent-to-data mapping is an RBF network, so flexibility is bounded by the basis, and the mapping is not invertible.

Normalizing flows~\citep{papamakarios,rezende,realnvp,brehmer,caterini2021rectangular} give exact likelihoods and invertible mappings but are not designed to produce a navigable low-dimensional map — their latent space has the same dimension as the data. Injective variants target density estimation on manifolds~\citep{brehmer} rather than visualisation. Some flows put the main structure in their first few latent coordinates~\citep{nested_dropout_flows, pcflows}, but they are built for density estimation, not 2-D visualisation.

This work combines GTM's fixed grid prior with a flow's invertibility and exact likelihood, exposing a scalar governing how much of the data's variance is forced into the visible map. 

Closest to \nfmap{} is FloDR~\citep{flodr}, developed independently and concurrently\footnote{Development of \nfmap{} began in May 2026; see the commit history at \url{https://github.com/hagax8/nfmap}.}. Both methods use a bijective normalizing flow whose leading coordinates form the 2D map while the remaining coordinates are retained as a residual, ensuring full invertibility and exact density evaluation. The two methods diverge in prior design and residual handling: \nfmap{} anchors the map prior to a fixed topographic grid and regulates the residual via a global scale $\sigma_\perp$, whereas FloDR fits an unconstrained Gaussian mixture to the embedding after training and models the residual conditionally to derive local spread diagnostics. Furthermore, while FloDR relies on an explicit ordinal ranking loss to preserve global ordering, \nfmap{} obtains global distance fidelity geometrically, through residual variance contraction. \nfmap{} additionally offers a Student-t mixture prior for large-scale map crowding.

We provide \nfmap{} as an open-source Python package available at \url{https://github.com/hagax8/nfmap}.

Contributions:

\begin{enumerate}
  \item The residual scale $\sigma_\perp$ as a control on the map, with a
        volume argument showing why, under a bijection, contracting
        $z_\perp$ forces structure into the visible coordinates $z_\parallel$.
  \item Empirical evidence that $\sigma_\perp$ trades local against global
        structure.
  \item A coupling analysis showing that $\sigma_\perp$, the grid extent, the repulsion weight and the graph weight all act on one quantity.
  \item The flow likelihood as an applicability-domain signal for projected data points.
  \item Mechanisms to address map crowding at dataset scale: a general heavy-tailed Student-t prior over the map coordinates, and a domain-specific hierarchical chemical kernel demonstrating the framework's extensibility to specialized data types.
\end{enumerate}

\begin{table}[H]
\centering
\small
\caption{Capabilities of each method.
\emph{Parametric}: unseen points can be projected onto an existing map.
\emph{Landscape grid}: the prior places $K$ components at fixed, known positions, so any quantity can be aggregated per node and rendered as a continuous background field at a resolution set by $K$. \emph{Student-t}: the low-dimensional
model uses Student-t tails, a standard remedy for crowding (built
into t-SNE's similarity kernel, and available in \nfmap{} as a Student-t mixture prior). UMAP's kernel, also used by FloDR, is heavy-tailed but not Student-t. }
\label{tab:capabilities}
\begin{tabular}{lccccc}
\toprule
 & Parametric & Invertible & Exact density & Landscape grid & Student-t \\
\midrule
t-SNE      & \ding{51}$^{*}$ & \ding{55} & \ding{55} & \ding{55} & \ding{51} \\
UMAP       & \ding{51}$^{*}$ & \ding{55} & \ding{55} & \ding{55} & \ding{55} \\
GTM        & \ding{51} & \ding{55} & \ding{51} & \ding{51} & \ding{55} \\
FloDR      & \ding{51} & \ding{51} & \ding{51} & \ding{55}$^{\dagger}$ & \ding{55} \\
\nfmap{}   & \ding{51} & \ding{51} & \ding{51} & \ding{51} & \ding{51}$^{\ddagger}$ \\
\bottomrule
\end{tabular}

\vspace{2pt}
\footnotesize $^{*}$via parametric variants (parametric t-SNE~\citep{parametric_tsne}, parametric UMAP~\citep{parametric_umap}); the standard implementations are non-parametric; $^{\dagger}$FloDR fits a Gaussian mixture to the embedding after training; its components are not fixed; $^{\ddagger}$optional
\end{table}

\newpage
\section{Methods}
\label{sec:methods}
\vspace{-0.3cm}

\subsection{Architecture}

Let $\mathbf{x} \in \mathbb{R}^{D}$ be an input observation. \nfmap{} first maps
the data to a PCA-reduced representation of dimension $d$ ($\mathbf{\tilde{x}} \in \mathbb{R}^d$), then learns a
\emph{bijection} $f_\theta : \mathbb{R}^{d} \to \mathbb{R}^{d}$ built from $B$
affine coupling blocks \citep{realnvp}. Block $b$ partitions its input
by a binary mask $\mathbf{m}_b$ and applies
\begin{equation}
  \mathbf{z}^{(b+1)} = \mathbf{z}^{(b)} \odot \exp\!\big(\mathbf{s}_b\big)
                     + \mathbf{t}_b ,
  \qquad
  (\mathbf{s}_b, \mathbf{t}_b) = (1-\mathbf{m}_b) \odot
      g_{\theta_b}\!\big(\mathbf{m}_b \odot \mathbf{z}^{(b)}\big),
  \label{eq:coupling}
\end{equation}
where $g_{\theta_b}$ is a two-hidden-layer GELU network and
$\mathbf{s}_b \!\leftarrow\! \tanh(\mathbf{s}_b)$ bounds the log-scale for
numerical stability. The final layer of each $g_{\theta_b}$ is zero-initialised,
so the flow begins as the identity. The log-determinant of the full map is the sum of the block terms,
$\log|\det \partial f_\theta / \partial \mathbf{\tilde{x}}| = \sum_b \mathbf{1}^{\!\top}\mathbf{s}_b$,
and the mapping is invertible in closed form.

The latent vector is partitioned into a two-dimensional \emph{map} and an
off-manifold \emph{residual},
\begin{equation}
  \mathbf{z} = f_\theta(\mathbf{\tilde{x}}) = (\mathbf{z}_\parallel, \mathbf{z}_\perp),
  \qquad
  \mathbf{z}_\parallel \in \mathbb{R}^{2}, \quad
  \mathbf{z}_\perp \in \mathbb{R}^{d-2}.
  \label{eq:split}
\end{equation}
Only $\mathbf{z}_\parallel$ is plotted; $\mathbf{z}_\perp$ is retained, so no
information is discarded and $f_\theta^{-1}$ reconstructs $\tilde{\mathbf{x}}$ exactly.

\subsection{Prior}

The prior factorises over the two blocks, combining a mixture model over the 2D map with a Gaussian penalty on the residual:
\begin{equation}
  p(\mathbf{z}) \;=\;
  p_{\mathrm{map}}(\mathbf{z}_\parallel)\;
  \mathcal{N}\!\big(\mathbf{z}_\perp;\, \mathbf{0},\, \sigma_\perp^{2}\mathbf{I}\big).
  \label{eq:prior}
\end{equation}
For the map prior $p_{\mathrm{map}}$, the standard construction is a Gaussian Mixture Model (GMM):
\begin{equation}
  p_{\mathrm{GMM}}(\mathbf{z}_\parallel) =
  \sum_{k=1}^{K} \pi_k\,
  \mathcal{N}\!\big(\mathbf{z}_\parallel;\, \boldsymbol{\mu}_k,\, s^{2}\mathbf{I}_2\big).
  \label{eq:prior_gmm}
\end{equation}
Alternatively, to address map crowding in larger datasets (where heavy-tailed distributions better accommodate moderate similarities), we replace the Gaussian components with Student-t distributions of $\nu$ degrees of freedom to form a Student-t Mixture Model (SMM):
\begin{equation}
  p_{\mathrm{SMM}}(\mathbf{z}_\parallel) =
  \sum_{k=1}^{K} \pi_k\,
  \mathcal{T}\!\big(\mathbf{z}_\parallel;\, \boldsymbol{\mu}_k,\, s^{2}\mathbf{I}_2,\, \nu\big).
  \label{eq:prior_smm}
\end{equation}
Following GTM~\citep{gtm}, in this paper, the component means
$\{\boldsymbol{\mu}_k\}$ are \emph{fixed} on a regular grid of radius $\texttt{grid\_bounds}$ and are not learned (although the option exists in \nfmap{}), and the mixing
weights are uniform, $\pi_k = 1/K$. Fixing $\pi_k$ matters: it denies the model
the option of explaining the data by reweighting which nodes are used, so the
only available mechanism is warping how points land across the grid. We use
$K=100$ and, for the Gaussian prior, $s = e^{-1} \approx 0.368$ throughout, a square grid for manifold datasets, and a hexagonal grid for others.

\subsection{Objective}

Training minimises
\begin{equation}
  \mathcal{L} \;=\;
  \underbrace{-\,\lambda_p \log p_{\mathrm{map}}(\mathbf{z}_\parallel)}_{\text{prior}}
  \;+\;
  \underbrace{\frac{1}{2\sigma_\perp^{2}}\,
     \frac{1}{d-2}\|\mathbf{z}_\perp\|^{2}}_{\text{manifold}}
  \;-\;
  \underbrace{\lambda_v \frac{2}{d}
     \log\!\Big|\det \tfrac{\partial f_\theta}{\partial \mathbf{\tilde{x}}}\Big|}_{\text{volume}}
  \;+\;
  \underbrace{\lambda_g \big(\mathcal{L}_{\mathrm{attr}}
            + \lambda_r \mathcal{L}_{\mathrm{rep}}\big)}_{\text{graph}}
  \;+\;
  \underbrace{\lambda_{\mathrm{inj}} \mathcal{L}_{\mathrm{inj}}}_{\text{injectivity}} .
  \label{eq:loss}
\end{equation}

The graph and injectivity terms act on $\mathbf{z}_\parallel$ only. Neighbour weights $w_{ij}$
come from UMAP's fuzzy simplicial set~\citep{umap} over the input space:
\begin{align}
  \mathcal{L}_{\mathrm{attr}}
    &= \mathbb{E}_{(i,j)\in E}\Big[\, w_{ij}\,
       \log\!\big(1 + \|\mathbf{z}_{\parallel i} - \mathbf{z}_{\parallel j}\|^{2}\big)\Big],
  \label{eq:attract}\\[2pt]
  \mathcal{L}_{\mathrm{rep}}
    &= \mathbb{E}_{(i,j)\sim \mathcal{U}}\left[\,
       \log\!\left(1 + \frac{1}{\|\mathbf{z}_{\parallel i}
         - \mathbf{z}_{\parallel j}\|^{2} + \epsilon}\right)\right],
  \label{eq:repel}\\[2pt]
  \mathcal{L}_{\mathrm{inj}}
    &= \mathbb{E}_{(i,j)\sim \mathcal{B}}\left[\,
       \exp\!\left(-\frac{\|\mathbf{z}_{\parallel i} - \mathbf{z}_{\parallel j}\|^{2}}{\tau^{2}}\right)
       \|\mathbf{\tilde{x}}_{i} - \mathbf{\tilde{x}}_{j}\|\right],
  \label{eq:inject}
\end{align}
where $(i,j) \sim \mathcal{U}$ are uniform non-neighbour pairs drawn from the batch, and $(i,j) \sim \mathcal{B}$ denotes all pairs within the mini-batch.
While repulsion acts as a global force preventing map collapse, the injectivity term is strictly local: by weighting a short-range Gaussian kernel ($\tau = 1.0$, fixed) by ambient distance, it penalises false neighbours without fighting true neighbours.

Eq.~\eqref{eq:loss} is a weighted objective rather than a likelihood. The manifold term uses the \emph{mean} over
residual coordinates rather than the sum a Gaussian negative log-likelihood
would give, which scales it by $1/(d-2)$; and the volume term is scaled by
$2/d$, which prevents the flow from inflating the log-determinant reward by stretching the $d-2$ residual dimensions without bound. These two choices largely compensate one another\footnote{Specifically, for an isotropic rescaling of the unscaled residual coordinates by a factor $a$ (such that $\mathbf{z}_\perp = a \mathbf{\tilde{x}}_\perp$), the exact likelihood is stationary at $a^{2} = \sigma_\perp^{2}(d-2)/\|\mathbf{\tilde{x}}_\perp\|^{2}$, and Eq.~\eqref{eq:loss} is stationary within a factor of two of that value.}. We use $\lambda_p = 0.1$, $\lambda_v = 1$, $\lambda_g = 5$, $\lambda_r = 1$, $\lambda_{\mathrm{inj}} = 0.5$, $\tau = 1.0$, $\epsilon = 10^{-4}$ ($\lambda_{\mathrm{inj}} = 0$ in the scale study).

\subsection{The $\sigma_\perp$ mechanism}
\label{sec:mechanism}

The residual term contributes gradient pressure on $\mathbf{z}_\perp$ scaling as
$\sigma_\perp^{-2}$. Because $f_\theta$ is bijective and the volume term
penalises contraction / collapse, variance removed from $\mathbf{z}_\perp$ cannot
vanish: it reappears in $\mathbf{z}_\parallel$. Decreasing $\sigma_\perp$
therefore forces more of the data's geometry into the two visible dimensions,
improving global distance fidelity at the cost of local neighbourhood fidelity.

Two consequences follow. First, the effect is bounded by how much variance there is to move: when the residual is low-dimensional there is no reservoir, and shrinking $\sigma_\perp$ instead destroys the dimension that legitimately carries the manifold's thickness. Second, if $\mathbf{z}_\parallel$ is bounded by a grid of fixed radius, the variance pushed into the map needs canvas to occupy, and the map may simply spill outside the grid boundaries. 

\subsection{Optimisation}

Parameters are fitted with AdamW (learning rate $10^{-3}$, linear warm-up followed by cosine decay) for
at most $1000$ epochs with early stopping on the training objective
(patience $100$; $300$ epochs and patience $30$ in the scale study). We use $B=6$ coupling blocks and a hidden width of $\max(64, \min(3d, 512))$.

\newpage

\section{Benchmark design}
\label{sec:benchmarks}
\subsection{Datasets}
We evaluate our method across three diverse dataset families. The first family comprises five chemical datasets: BACE, BBBP, and ESOL from MoleculeNet~\citep{moleculenet}, alongside ChEMBL~\citep{chembl} activity sets for DRD2 and EGFR. Molecules are represented by 2048-bit Morgan fingerprints ($r=2$) concatenated with eight standard physicochemical descriptors (computed via RDKit~\citep{rdkit}). High-dimensional features are standardised, reduced to $d=32$ principal components (or $d=128$ for the scale study), and standardised again, while low-dimensional benchmarks retain their original dimension $d=D$. 
The second family consists of standard low-dimensional machine learning benchmarks: Digits, Wine, Breast Cancer, Iris, Two Moons, and Concentric Circles. The third family comprises synthetic 3-D manifolds designed to test geometric unfolding: the Swiss Roll, the S-Curve, and a synthetic Severed Sphere. 
With the exception of the chemical datasets and the severed sphere, all standard benchmarks and manifolds are generated or sourced using \texttt{scikit-learn}~\citep{scikit-learn}. Experiments use a 600-sample subset of each dataset, each configuration is fitted across three random seeds (five in the prior ablation), and we report the mean metric.
A separate scale study (Section~\ref{sec:scale}) uses larger full-scale ChEMBL activity sets: DRD2 (869 molecules), hERG (9{,}400) and EGFR (10{,}573), because the crowding effects it examines are properties of large maps and are invisible at the subsample size used above. For that experiment, we use 128 principal components instead of 32, which is closer to what would be used in production cheminformatics pipelines.

\subsection{Baselines and Implementation}
For our baseline comparisons, we use the \texttt{scikit-learn}~\citep{scikit-learn} implementation of t-SNE. For UMAP, we use the official \texttt{umap-learn}~\citep{umap} Python package. Finally, for Generative Topographic Mapping (GTM), we use the \texttt{ugtm} package~\citep{gaspar2018ugtm} implementation. All baseline models receive the exact same PCA-reduced input matrices as \nfmap{}.

\subsection{Metrics}

Let $\mathbf{x}_i$ denote inputs and $\mathbf{z}_i \in \mathbb{R}^2$ their map
coordinates, with $d^X_{ij}$ and $d^Z_{ij}$ the corresponding pairwise
distances.

\paragraph{Global structure.} Distance correlation is the Pearson correlation
between the two distance vectors,
\begin{equation}
  \mathrm{DistCor} \;=\;
  \frac{\sum_{i<j}\big(d^X_{ij}-\bar{d}^X\big)\big(d^Z_{ij}-\bar{d}^Z\big)}
       {\sqrt{\sum_{i<j}\big(d^X_{ij}-\bar{d}^X\big)^2}
        \sqrt{\sum_{i<j}\big(d^Z_{ij}-\bar{d}^Z\big)^2}} .
  \label{eq:distcor}
\end{equation}

\paragraph{Local structure.} We use trustworthiness and continuity
\citep{Venna2005LocalMS}. With $\mathcal{N}^X_k(i)$ and $\mathcal{N}^Z_k(i)$ the
$k$-neighbourhoods of $i$ in the two spaces and $r(i,j)$ the rank of $j$ among
$i$'s input-space neighbours,
\begin{equation}
  T(k) = 1 - \frac{2}{nk(2n-3k-1)}
    \sum_{i=1}^{n}\sum_{j \in \mathcal{N}^Z_k(i)\setminus \mathcal{N}^X_k(i)}
    \big(r(i,j)-k\big),
  \label{eq:trust}
\end{equation}
with continuity $C(k)$ defined symmetrically by exchanging the two spaces. We
report their harmonic mean, $\mathrm{T/C\ F1} = 2TC/(T+C)$, at $k=5$. 

\paragraph{Geodesic structure.} On a manifold that folds back on itself, Eq.~\eqref{eq:distcor} compares against \emph{ambient} distances and therefore rewards a map that leaves the manifold rolled. For the synthetic manifold datasets we build a $k$-nearest-neighbour graph ($k=10$) on the
input and take all-pairs shortest paths $g_{ij}$ \citep{isomap},
reporting $\mathrm{DistCor}_{\mathrm{geo}}$ as Eq.~\eqref{eq:distcor} with
$d^X_{ij}$ replaced by $g_{ij}$.

\paragraph{Chemical structure.} For chemistry we also measure whether molecules sharing a Bemis--Murcko scaffold remain together. With $c_i$ the scaffold identifier of
molecule $i$ and $\mathcal{E} = \{i : |\{j \neq i : c_j = c_i\}| \ge 1\}$ the
molecules whose scaffold has at least one other member,
\begin{equation}
  \mathrm{purity}@k \;=\;
  \frac{1}{|\mathcal{E}|}\sum_{i \in \mathcal{E}}
  \frac{1}{k}\Big|\big\{\, j \in \mathcal{N}^Z_k(i) \;:\; c_j = c_i \,\big\}\Big| .
  \label{eq:purity}
\end{equation}
Singleton scaffolds are excluded: they can never have a same-scaffold neighbour. We report purity against the chance rate
$\frac{1}{|\mathcal{E}|}\sum_{i\in\mathcal{E}}(|c_i|-1)/(n-1)$. We also report the silhouette coefficient over map coordinates with $c_i$ as the label, restricted to the same $\mathcal{E}$, which captures separation between scaffold groups rather than only neighbourhood composition.

\paragraph{Density as an applicability domain.} \nfmap{} yields an exact data density
$\log p(\mathbf{x})$. We assess it as an out-of-distribution score by the area under the ROC curve separating held-out in-distribution points from out-of-distribution ones, against two baselines: mean distance to the $k$ nearest training points, and PCA reconstruction error.
\subsection{Experiments}
\begin{enumerate}
  \item \textbf{$\sigma_\perp$ as a local/global control}: Examining the effect of the residual variance parameter across different dataset geometries.
  \item \textbf{Graph loss weights}: Examining how the attraction and repulsion weights provide further control on map spread.
  \item \textbf{Joint $\sigma_\perp \times$ \texttt{grid\_bounds} sweep}: Sweeping eight values of $\sigma_\perp$ spanning four orders of magnitude against three grid radii (6, 12, 24) on the chemical datasets to reveal parameter coupling (Figure~\ref{fig:joint}).
  \item \textbf{Comparison at matched tuning}: Each method is swept over its own local/global control (14 values of $\sigma_\perp$ from $10^{-3}$ to $10^{3}$ with \texttt{grid\_bounds} 12 and 24 for \nfmap{}, perplexity for t-SNE ($5$--$500$), \texttt{n\_neighbors} and \texttt{min\_dist} for UMAP) and compared at its best setting.
  \item \textbf{Prior ablation}: Testing a single Gaussian, a learned mixture, and our fixed grid to measure the cost of the structural constraint.
  \item \textbf{Applicability domain}: Up to three out-of-distribution constructions (feature-permuted inputs and matched-moment Gaussian noise for every dataset; held-out novel classes where there are at least three classes) across fourteen datasets using exact likelihood.
  \item \textbf{Scale study and domain extensions}: Evaluating the hierarchical chemical kernel and Student-t prior on larger DRD2, hERG and EGFR ChEMBL datasets, including a degrees-of-freedom sweep.
\end{enumerate}

\newpage
\section{Results}
\label{sec:results}
\subsection{\texorpdfstring{$\sigma_\perp$}{sigma-perp} is a local/global control whose direction depends on the data}
\begin{figure}[H]
  \centering
  \includegraphics[width=0.75\linewidth]{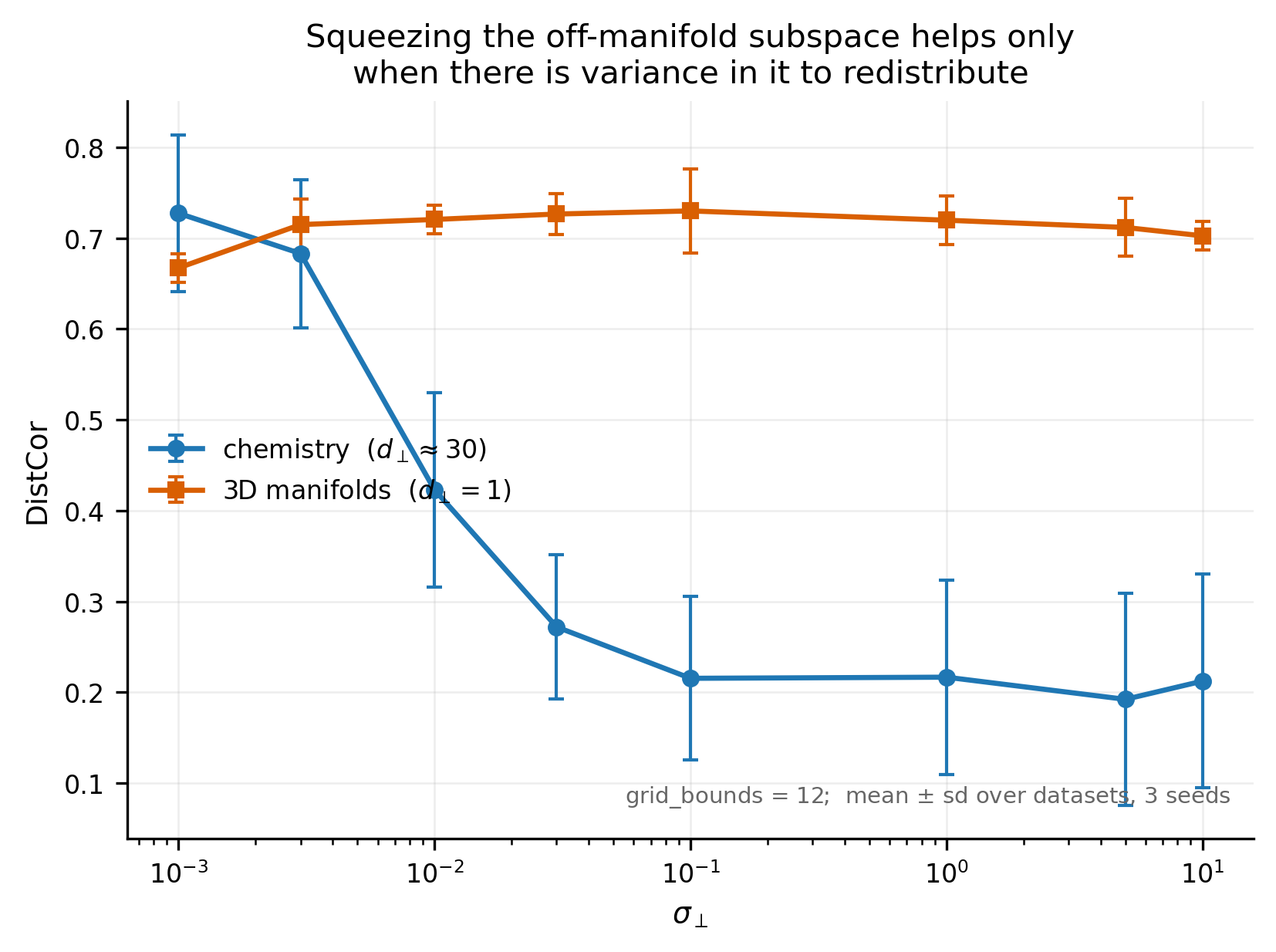}
  \caption{The direction of the $\sigma_\perp$ trade-off is set by the data's geometry. Mean $\pm$ s.d. over datasets and three seeds, \texttt{grid\_bounds}$=12$.}
  \label{fig:regimes}
\end{figure}

Shrinking $\sigma_\perp$ transfers variance into the map, improving global distance fidelity at the cost of neighbourhood fidelity; but only where the residual has variance to give. Figure~\ref{fig:regimes} shows the two regimes on one axis: on chemistry ($d_\perp \approx 30$) DistCor falls from $0.73$ to $0.21$ as $\sigma_\perp$ rises, whereas on the 3-D manifolds ($d_\perp = 1$) it rises from $0.67$ and flattens near $0.73$. The curves cross. With one off-manifold dimension there is no reservoir to redistribute, and shrinking $\sigma_\perp$ instead destroys the dimension carrying the manifold's thickness.

\newpage
\subsection{The graph loss terms (attraction and repulsion) control map spread}

\begin{figure}[H]
  \centering
  \includegraphics[width=\linewidth]{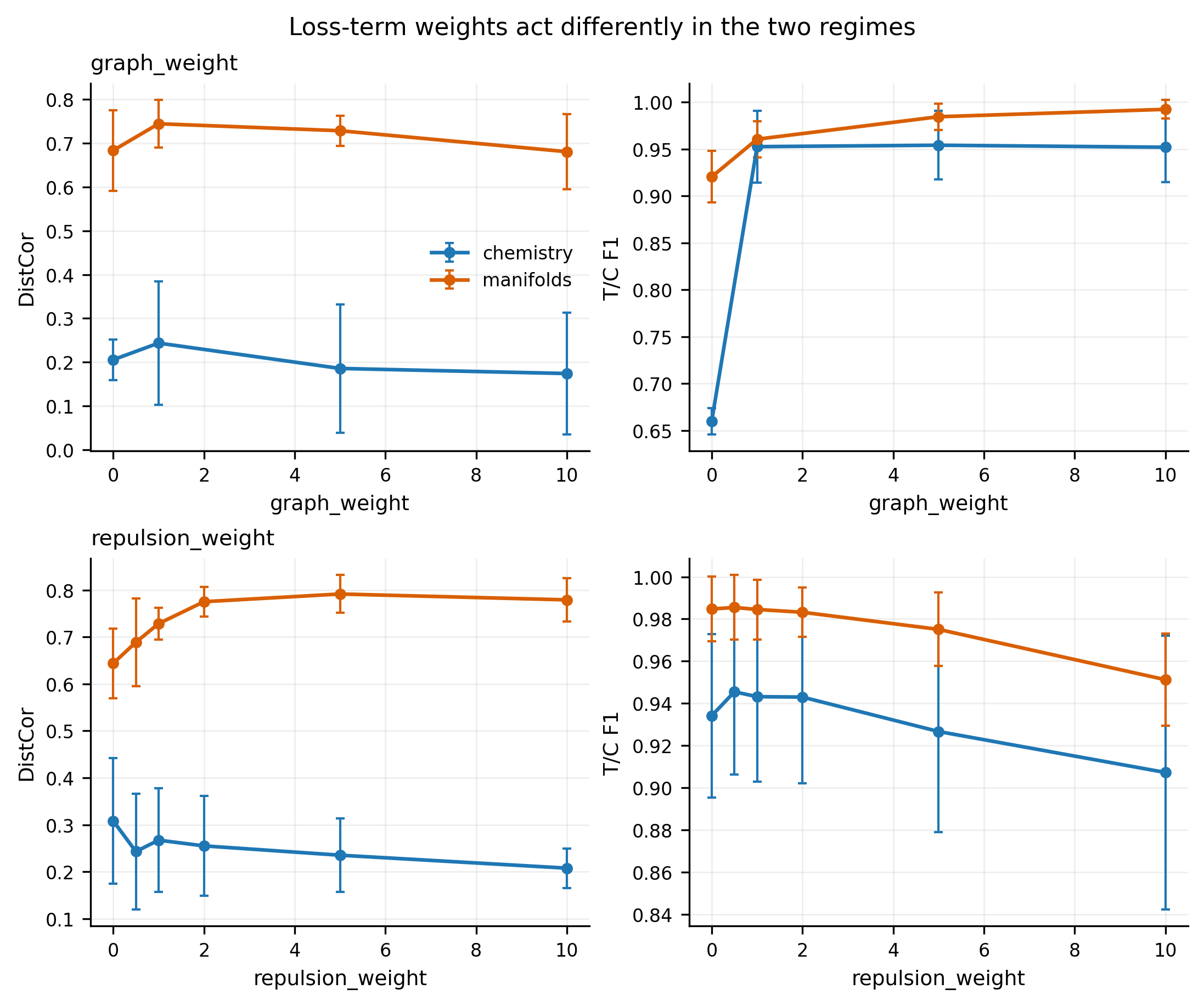}
  \caption{Ablation of the graph attraction ($\lambda_g$) and repulsion ($\lambda_r$) weights. The graph terms act as opposing forces on the map's spread relative to the canvas.}
  \label{fig:loss_weights}
\end{figure}

In addition to $\sigma_\perp$ and the grid radius, the graph loss weights provide further control over embedding geometry (Figure~\ref{fig:loss_weights}). 

The attraction term is strictly necessary to produce a usable map: at $\lambda_g = 0$, local neighbourhood fidelity (T/C F1) drops for all dataset families, indicating that the prior and volume terms alone are insufficient to preserve structure. 

Both graph terms act as additional handles on the same underlying quantity: the spread of the 2-D coordinates relative to the fixed grid canvas. Stronger attraction ($\lambda_g$) contracts the map, effectively halving the fraction of out-of-bounds points on manifold datasets. Conversely, the repulsion term ($\lambda_r$) acts as an anti-collapse mechanism that expands the map.

\newpage
\subsection{\texorpdfstring{$\sigma_\perp$}{sigma-perp} cannot be chosen independently of the grid}
\begin{figure}[H]
  \centering
  \includegraphics[width=\linewidth]{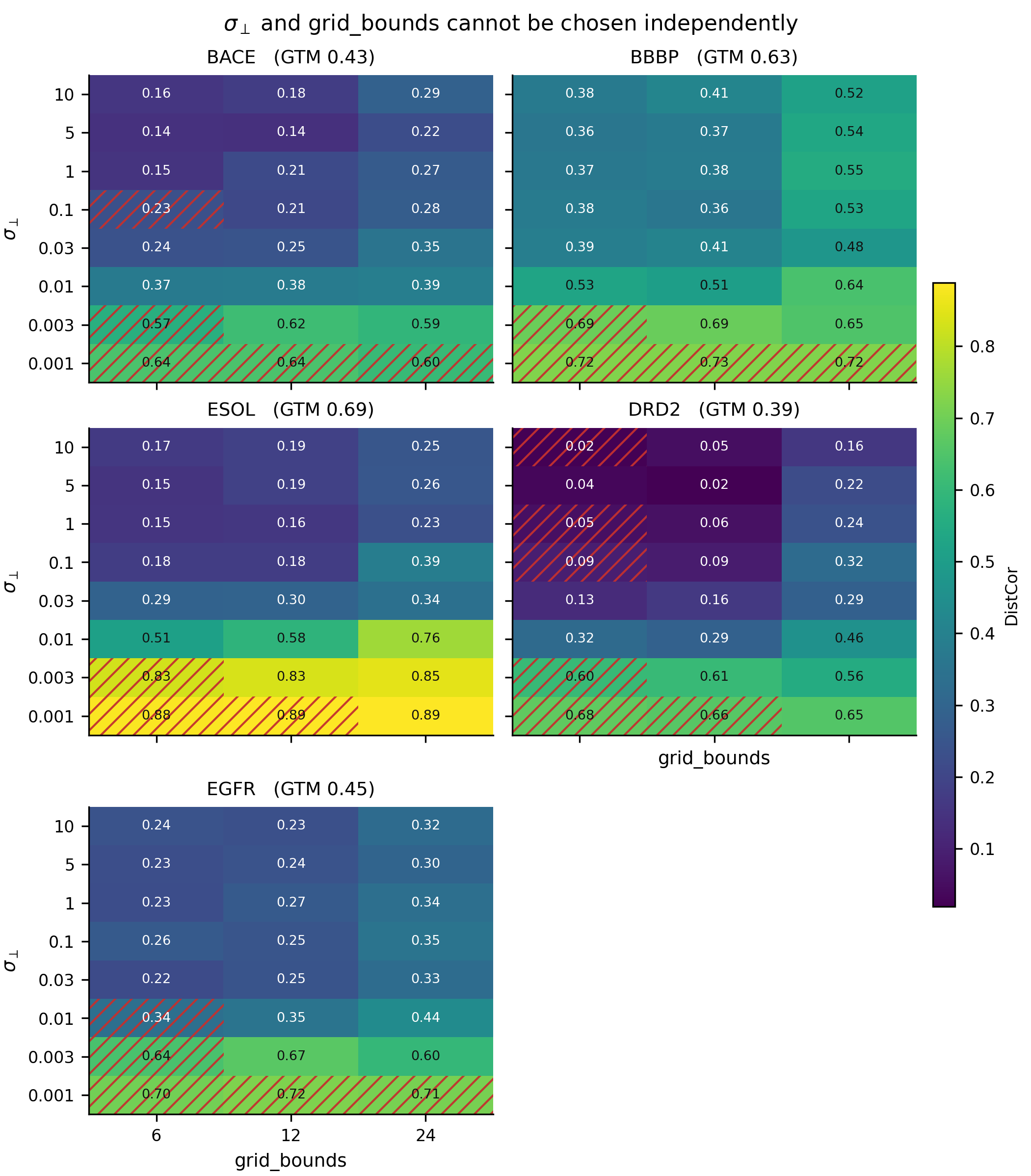}
  \caption{Joint sweep on the chemical datasets. Hatched cells violate the containment constraint.}
  \label{fig:joint}
\end{figure}

Figure~\ref{fig:joint} sweeps $\sigma_\perp$ and \texttt{grid\_bounds} together. Widening the canvas from 6 to 12 allows $\sigma_\perp$ an order of magnitude lower (for high-dimensional datasets); the wider grid absorbs the redistributed variance. In this example, the best shared ``global'' configuration (optimizing DistCor) keeping $\ge 95\%$ of points inside the grid is $\sigma_\perp = 0.003$, \texttt{grid\_bounds} $=12$. The gain saturates: widening further to \texttt{grid\_bounds}$\,=24$ does not admit $\sigma_\perp = 0.001$ within the containment constraint and lowers mean DistCor slightly ($0.65$ vs.\ $0.68$), so the shared optimum sits at the intermediate grid. 

\newpage
\subsection{Comparison with existing methods}

\begin{table}[H]
\centering
\footnotesize
\setlength{\tabcolsep}{3.5pt}
\caption{Each method at its own best setting (means over 3 seeds, CPU). G: global structure (DistCor; geodesic DistCor on the 3-D manifolds). L: local structure (T/C F1). Each sweep is reduced to its Pareto front over (G, L); the reported point is the front's knee, or its maximum-G point on the manifolds. Best per dataset in bold. \nfmap{}'s reported maps keep at least 99\% of points inside \texttt{grid\_bounds}, except the S-curve (94\%). Settings in Table~\ref{tab:settings}.}
\label{tab:comparison}
\begin{tabular}{lcccccccc}
\toprule
 & \multicolumn{2}{c}{\nfmap{}} & \multicolumn{2}{c}{GTM} & \multicolumn{2}{c}{t-SNE} & \multicolumn{2}{c}{UMAP} \\
\cmidrule(lr){2-3}\cmidrule(lr){4-5}\cmidrule(lr){6-7}\cmidrule(lr){8-9}
 & G & L & G & L & G & L & G & L \\
\midrule
\multicolumn{9}{l}{\textit{Chemistry}} \\
\quad BACE & 0.37 & 0.97 & 0.34 & 0.96 & \textbf{0.61} & \textbf{0.98} & 0.30 & \textbf{0.98} \\
\quad BBBP & 0.55 & 0.89 & \textbf{0.71} & 0.87 & 0.59 & 0.91 & 0.39 & \textbf{0.92} \\
\quad ESOL & \textbf{0.76} & 0.85 & 0.70 & 0.87 & 0.57 & 0.89 & 0.18 & \textbf{0.91} \\
\quad DRD2 & 0.36 & 0.96 & 0.42 & 0.95 & \textbf{0.59} & 0.96 & 0.23 & \textbf{0.98} \\
\quad EGFR & 0.44 & 0.92 & 0.47 & 0.91 & \textbf{0.67} & 0.92 & 0.34 & \textbf{0.95} \\
\multicolumn{9}{l}{\textit{3-D manifolds}} \\
\quad Swiss roll & 0.52 & 0.99 & 0.26 & 0.98 & 0.77 & \textbf{1.00} & \textbf{0.86} & \textbf{1.00} \\
\quad S-curve & 0.94 & 0.99 & 0.50 & 0.99 & \textbf{0.99} & \textbf{1.00} & \textbf{0.99} & \textbf{1.00} \\
\quad Severed sphere & 0.92 & \textbf{1.00} & 0.91 & 0.99 & \textbf{0.95} & \textbf{1.00} & \textbf{0.95} & \textbf{1.00} \\
\multicolumn{9}{l}{\textit{Other}} \\
\quad Digits & 0.63 & \textbf{0.97} & 0.53 & \textbf{0.97} & \textbf{0.70} & 0.96 & 0.59 & 0.96 \\
\quad Iris & \textbf{0.98} & \textbf{0.98} & 0.75 & 0.95 & \textbf{0.98} & \textbf{0.98} & 0.91 & \textbf{0.98} \\
\quad Cancer & 0.80 & 0.93 & 0.66 & 0.94 & \textbf{0.81} & 0.95 & 0.74 & \textbf{0.96} \\
\quad Wine & 0.84 & 0.94 & 0.72 & 0.94 & \textbf{0.86} & 0.95 & 0.78 & \textbf{0.96} \\
\bottomrule
\end{tabular}
\end{table}

On global structure, t-SNE at high perplexity is the strongest baseline on most datasets. \nfmap{} has the best global fidelity on ESOL, ties t-SNE on Iris, and is above UMAP on all five chemical sets, while its local fidelity stays within 0.06 of the best method throughout. It trails t-SNE and UMAP on the geodesic structure of the Swiss roll. Unlike either, it does this with an invertible map. For t-SNE, the global-structure optimum lies at a perplexity approaching the sample size, which is impractical at realistic $n$.

\begin{figure}[H]
  \centering
  \includegraphics[width=\linewidth]{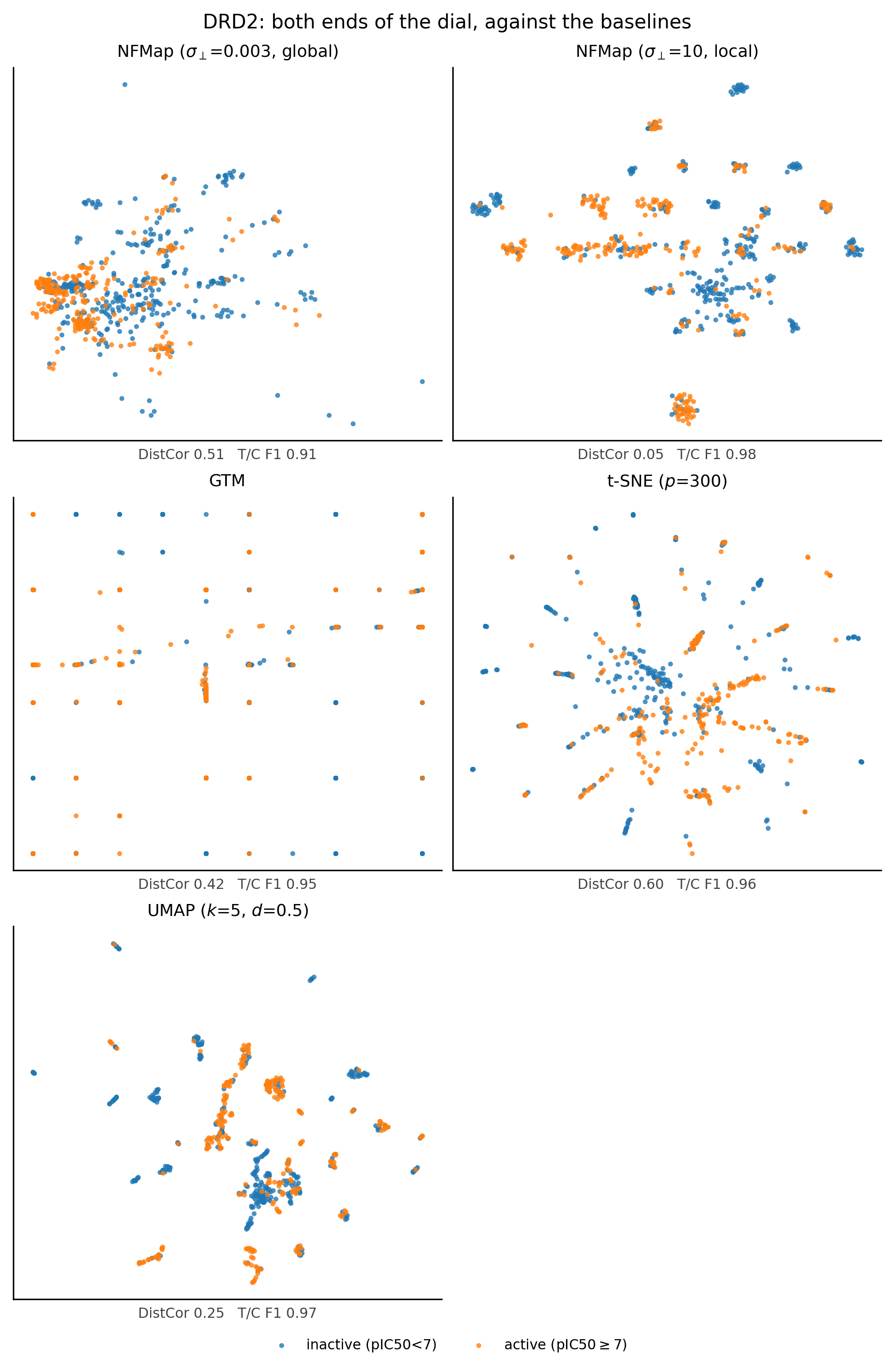}
  \caption{DRD2 at both ends of the $\sigma_\perp$ dial, against GTM, t-SNE and UMAP.}
  \label{fig:maps_drd2}
\end{figure}

\begin{figure}[H]
  \centering
  \includegraphics[width=\linewidth]{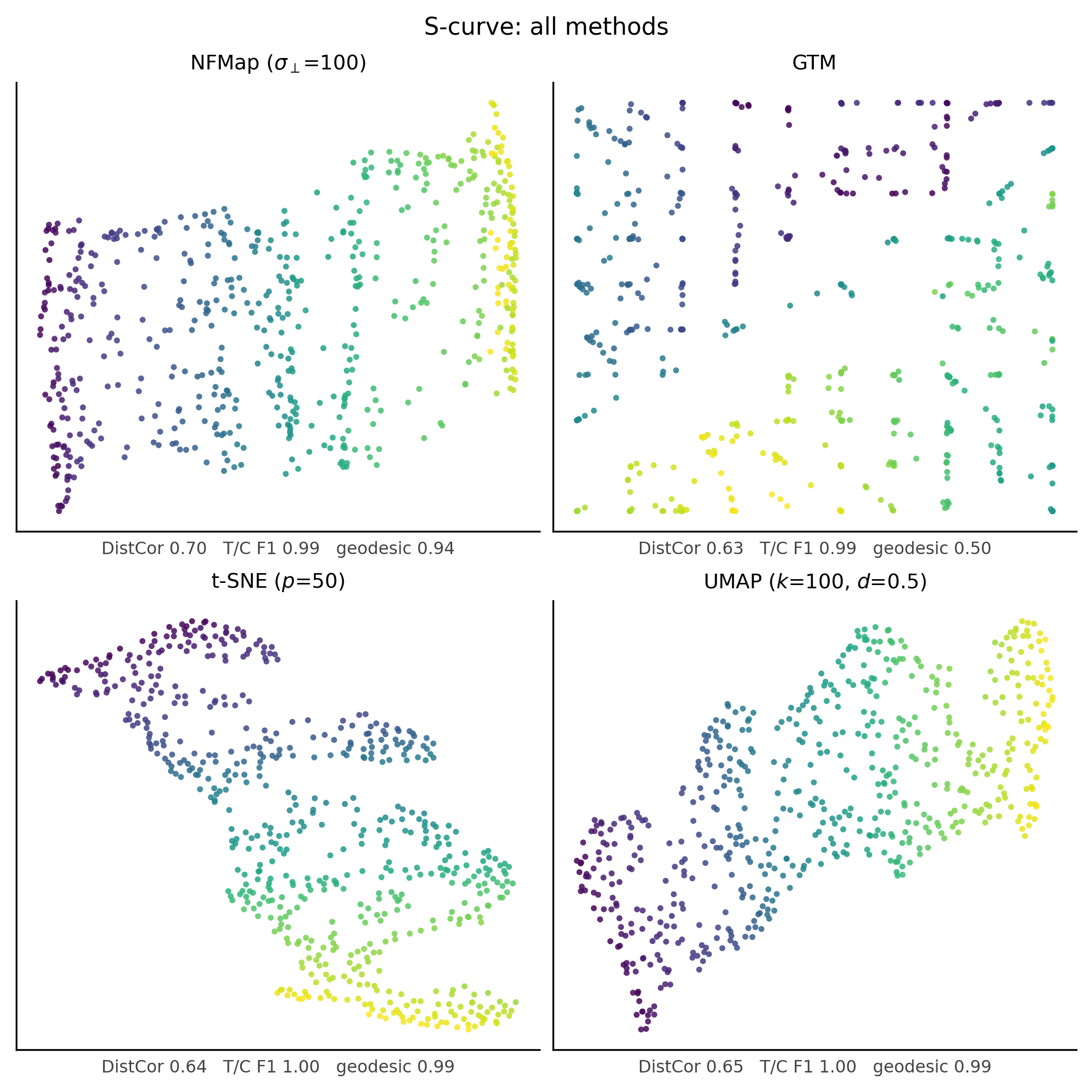}
  \caption{S-curve. \nfmap{} and baselines at their Table~\ref{tab:comparison} settings.}
  \label{fig:maps_s}
\end{figure}

Figures~\ref{fig:maps_drd2} and \ref{fig:maps_s} show maps for DRD2 and the S-curve. For S-curve, we used the best geodesic setup for every map; for DRD2, we show two extreme settings for \nfmap{}, with $\sigma_\perp$ = 0.003 (global-optimizing) and $\sigma_\perp$ = 10 (local-optimizing).

\newpage

\newpage
\subsection{The fixed grid does not harm performance}
\begin{figure}[H]
  \centering
  \includegraphics[width=\linewidth]{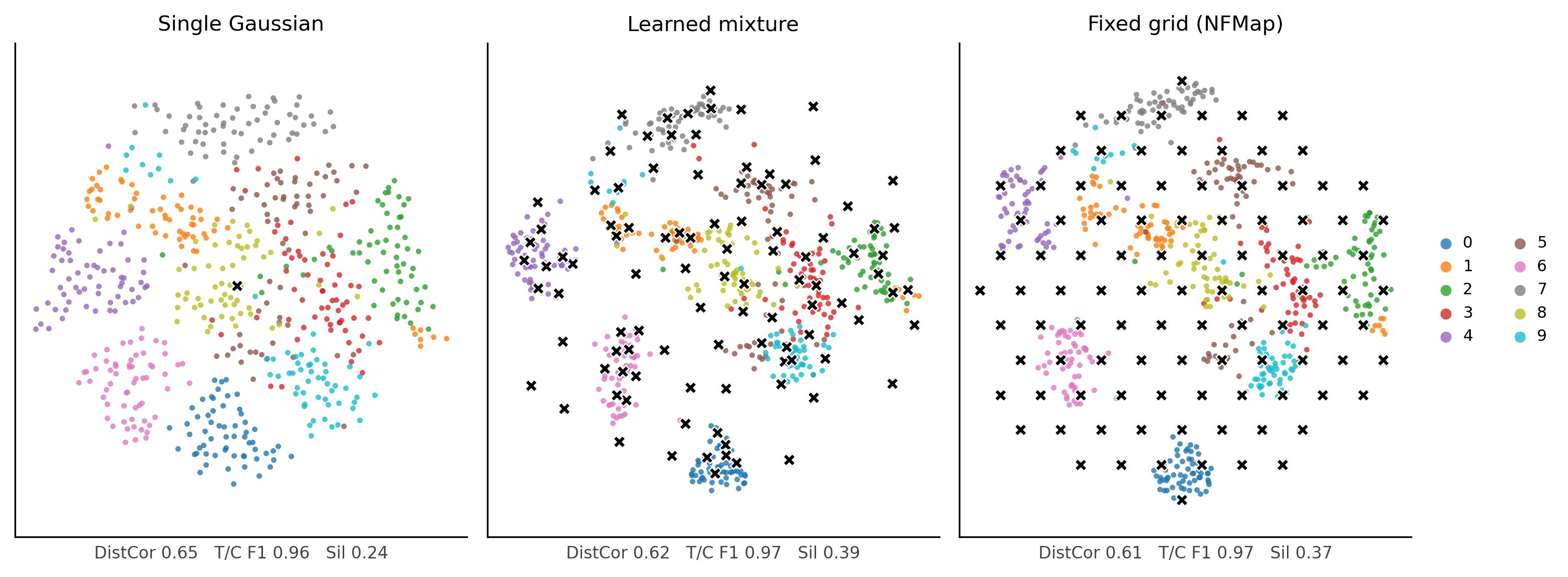}
  \caption{The prior over the \nfmap{} coordinates, on digits ($n=600$, 64 features, coloured by digit). Left: a single Gaussian. Middle: a mixture of 100 components whose means are learned. Right: the same mixture with its means pinned to a fixed hexagonal grid. Crosses mark the mixture means after training: the learned means leave the lattice and gather in the clusters, while the fixed ones stay where they were placed. All three panels at grid\_bounds $=6$ and $\sigma_\perp=10$. DistCor, T/C F1 and silhouette under each panel are means over 3 seeds.}
  \label{fig:prior}
\end{figure}

Fixing the grid (Table~\ref{tab:prior}, Figure~\ref{fig:prior}) is statistically non-inferior to learning the mixture means, for both DistCor (margin $0.02$; one-sided paired $t$-test, $p = 8\times10^{-6}$) and T/C F1 (margin $0.01$; $p = 9\times10^{-11}$), over 43 paired runs on 9 datasets.
We therefore adopt the fixed grid for the canonical grid cells it provides.

\begin{table}[H]
\centering
\small
\caption{Prior ablation across datasets and seeds (mean $\pm$ sd; 3 seeds for Digits, 5 seeds for the other datasets). Against a single Gaussian the fixed grid gives equal or better T/C F1 and trades some DistCor on Digits and the 3-D manifolds. Off grid: share of points outside grid\_bounds (also an average across datasets and seeds). A single Gaussian has no grid, so off-grid is not reported.}
\label{tab:prior}
\begin{tabular}{lccc}
\toprule
Prior & DistCor & T/C F1 & Off grid \\
\midrule
\multicolumn{4}{l}{\textit{Digits}} \\
\quad Single Gaussian & $0.648 \pm 0.002$ & $0.963 \pm 0.004$ & -- \\
\quad Learned mixture & $0.624 \pm 0.005$ & $0.973 \pm 0.002$ & 0.0\% \\
\quad Fixed grid (\nfmap{}) & $0.608 \pm 0.010$ & $0.973 \pm 0.003$ & 0.9\% \\
\addlinespace
\multicolumn{4}{l}{\textit{Chemistry (5 datasets)}} \\
\quad Single Gaussian & $0.657 \pm 0.090$ & $0.848 \pm 0.049$ & -- \\
\quad Learned mixture & $0.679 \pm 0.086$ & $0.856 \pm 0.050$ & 12.0\% \\
\quad Fixed grid (\nfmap{}) & $0.677 \pm 0.095$ & $0.858 \pm 0.050$ & 3.0\% \\
\addlinespace
\multicolumn{4}{l}{\textit{3-D manifolds (3 datasets)}} \\
\quad Single Gaussian & $0.756 \pm 0.041$ & $0.940 \pm 0.024$ & -- \\
\quad Learned mixture & $0.682 \pm 0.048$ & $0.992 \pm 0.010$ & 3.1\% \\
\quad Fixed grid (\nfmap{}) & $0.719 \pm 0.028$ & $0.989 \pm 0.012$ & 1.7\% \\
\addlinespace
\bottomrule
\end{tabular}
\end{table}

\newpage

\subsection{The likelihood is an applicability-domain signal}
\begin{figure}[H]
  \centering
  \includegraphics[width=\linewidth]{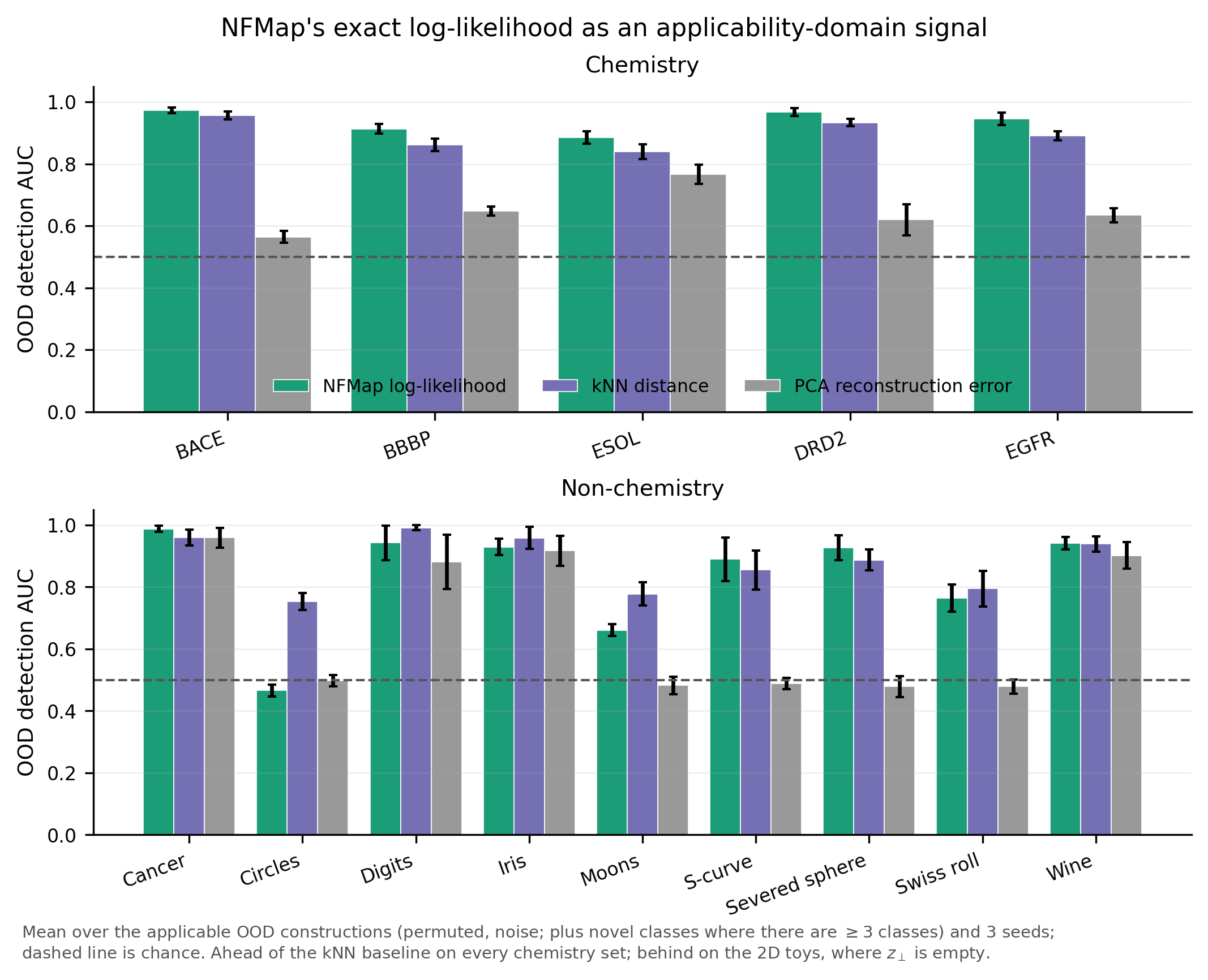}
  \caption{Out-of-distribution detection AUC, averaged over the applicable OOD constructions (feature-permuted inputs and matched-moment noise for every dataset; held-out novel classes in addition where there are $\geq$3 classes) and three seeds. Dashed line is chance.}
  \label{fig:ad}
\end{figure}

Figure~\ref{fig:ad} reports out-of-distribution AUC. \nfmap{}'s exact log-likelihood leads the $k$-NN distance baseline on every chemical dataset (mean $0.937$ versus $0.896$). On the non-chemical sets it trails on average ($0.834$ versus $0.880$), most clearly on the two-dimensional toys (circles, moons), where $d_\perp = 0$ and the density has no residual to measure; results on the remaining sets are mixed (ahead on wine, cancer, S-curve and severed sphere; behind on digits, iris and swiss roll).

\subsection{Crowding: effects that appear only at scale}
\label{sec:scale}
\begin{figure}[H]
  \centering
  \includegraphics[width=\linewidth]{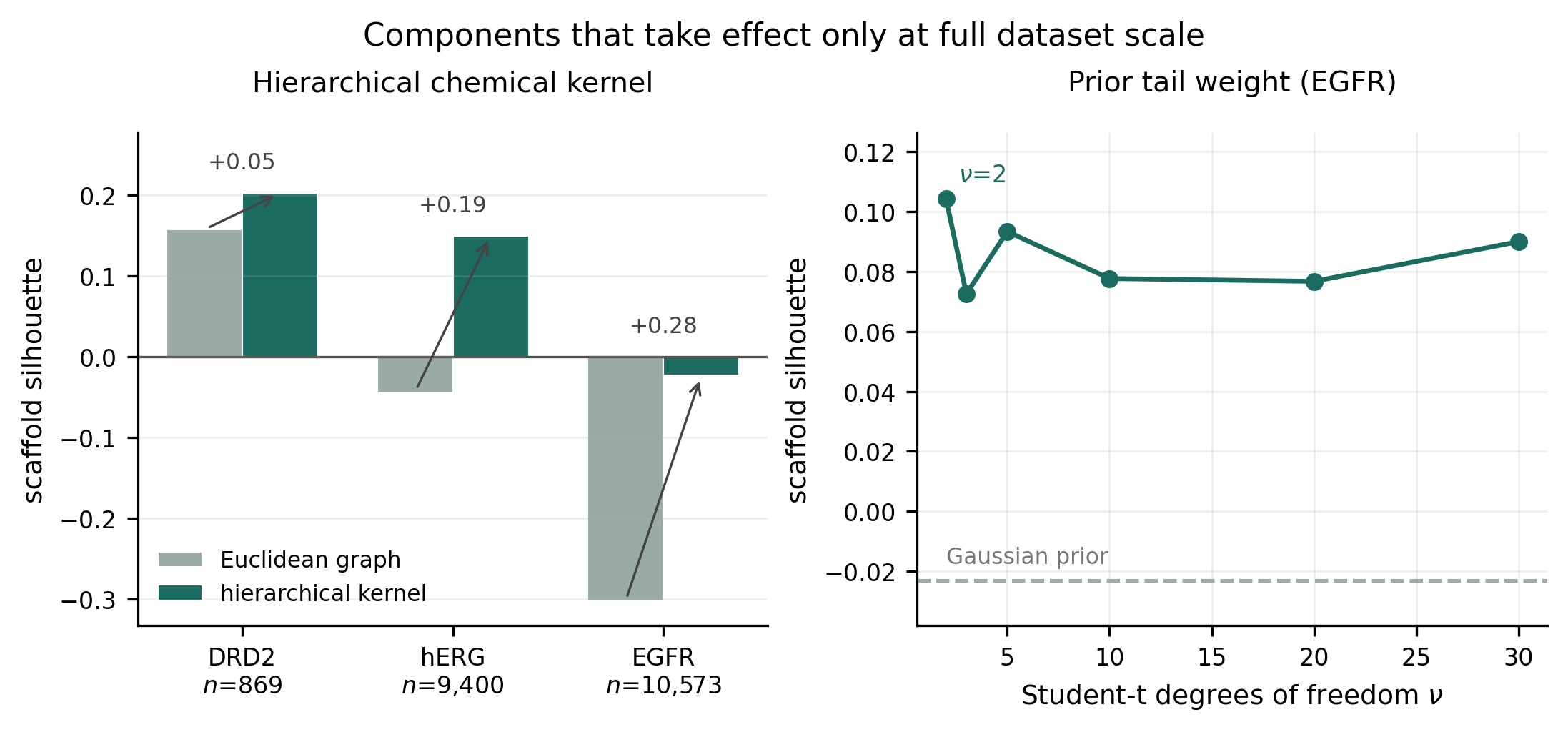}
  \caption{Crowding remedies at full dataset scale (single seed). Left: replacing the Euclidean neighbourhood graph with the hierarchical chemical kernel, on the complete DRD2, hERG and EGFR activity sets. Right: the Student-t degrees-of-freedom sweep on EGFR, with the Gaussian prior shown for reference.}
  \label{fig:scale}
\end{figure}

The experiments so far use small datasets (at most $n=600$), where crowding is not addressed: in high dimensions, there is far more room at moderate distances than two dimensions can represent, so as $n$ grows the moderately dissimilar points are increasingly compressed together. At $n=600$ there is little crowding to relieve. We investigated the crowding problem in larger chemical datasets, by adding two features: a scaffold-informed hierarchical chemical kernel (Bemis--Murcko scaffold and generic skeleton), and a Student-t prior over the map coordinates (Figure~\ref{fig:scale}). 
Replacing the Euclidean graph with the hierarchical kernel raises scaffold silhouette by $+0.045$ on DRD2 ($n=869$), $+0.194$ on hERG ($n=9{,}400$) and $+0.280$ on EGFR ($n=10{,}573$). On EGFR, a Student-t prior over the map coordinates further raises scaffold purity from $0.571$ to $0.63$–$0.64$ and silhouette from $-0.023$ to between $+0.07$ and $+0.10$, with little dependence on the degrees of freedom across $\nu = 2$–$30$. The best EGFR configuration (hierarchical kernel, Student-t, $\nu=2$) reaches scaffold purity $0.642$ and silhouette $+0.104$, against $0.388$ and $-0.303$ for the Euclidean-graph Gaussian baseline.

\subsection{Discussion}
Treating the 2-D map as one half of a bijection, with a volume term that penalises contraction, gives $\sigma_\perp$ a concrete meaning: it sets how much of the data's variance the flow may leave in the residual. On high-dimensional data, shrinking it moves geometry onto the map and improves global structure; on thin 3-D manifolds it flattens the dimension that carries the manifold's thickness. The right setting therefore depends on the data's geometry rather than being a universal default.

The coupling between $\sigma_\perp$ and the grid extent also means that grid-based methods should not be tuned one parameter at a time. Chosen jointly, the grid absorbs the redistributed variance, and \nfmap{} reaches global fidelity close to GTM's and above UMAP's on the chemical datasets without spilling points off the canvas. Fixing the grid, rather than learning the mixture means, had no measurable cost in embedding quality.

The flow formulation also gives \nfmap{} capabilities that neighbour embeddings lack. Its exact likelihood separates in- from out-of-distribution molecules better than a $k$-NN distance baseline on all five chemical datasets. At larger scale (up to $\sim$10{,}000 molecules), where crowding becomes visible, a Student-t prior and a scaffold-aware kernel each improve scaffold separation, and both fit into the framework without changing the flow.

\section{Conclusions}
\vspace{-0.3cm}

We introduced \nfmap{}, an invertible, parametric dimensionality reduction method that combines GTM's fixed latent grid with the exact likelihood of a normalizing flow.
The residual scale $\sigma_\perp$ controls the local/global trade-off; the direction of its effect depends on the intrinsic dimensionality of the data, and it must be chosen together with the grid extent.
Unlike t-SNE and UMAP, \nfmap{} is fully invertible. It also projects new points onto a trained map, carries any map position back to data space, and its density works as an out-of-distribution signal on chemical data. The fixed grid provides a common reference canvas at no measurable cost in embedding quality.
The framework is also easy to extend: a Student-t prior and a hierarchical chemical kernel improve map structure on larger datasets.
The main limitations are that several hyperparameters affect how far the map spreads, and that points can fall outside the grid. Adaptive calibration of the grid extent is a natural next step.

\small
\bibliographystyle{unsrtnat}
\bibliography{main}
\newpage
\appendix
\counterwithin{figure}{section}
\counterwithin{table}{section}

\section{Additional Qualitative Results on 3-D Manifolds}
\label{app:manifold_maps}

\begin{figure}[H]
  \centering
  \includegraphics[width=\linewidth]{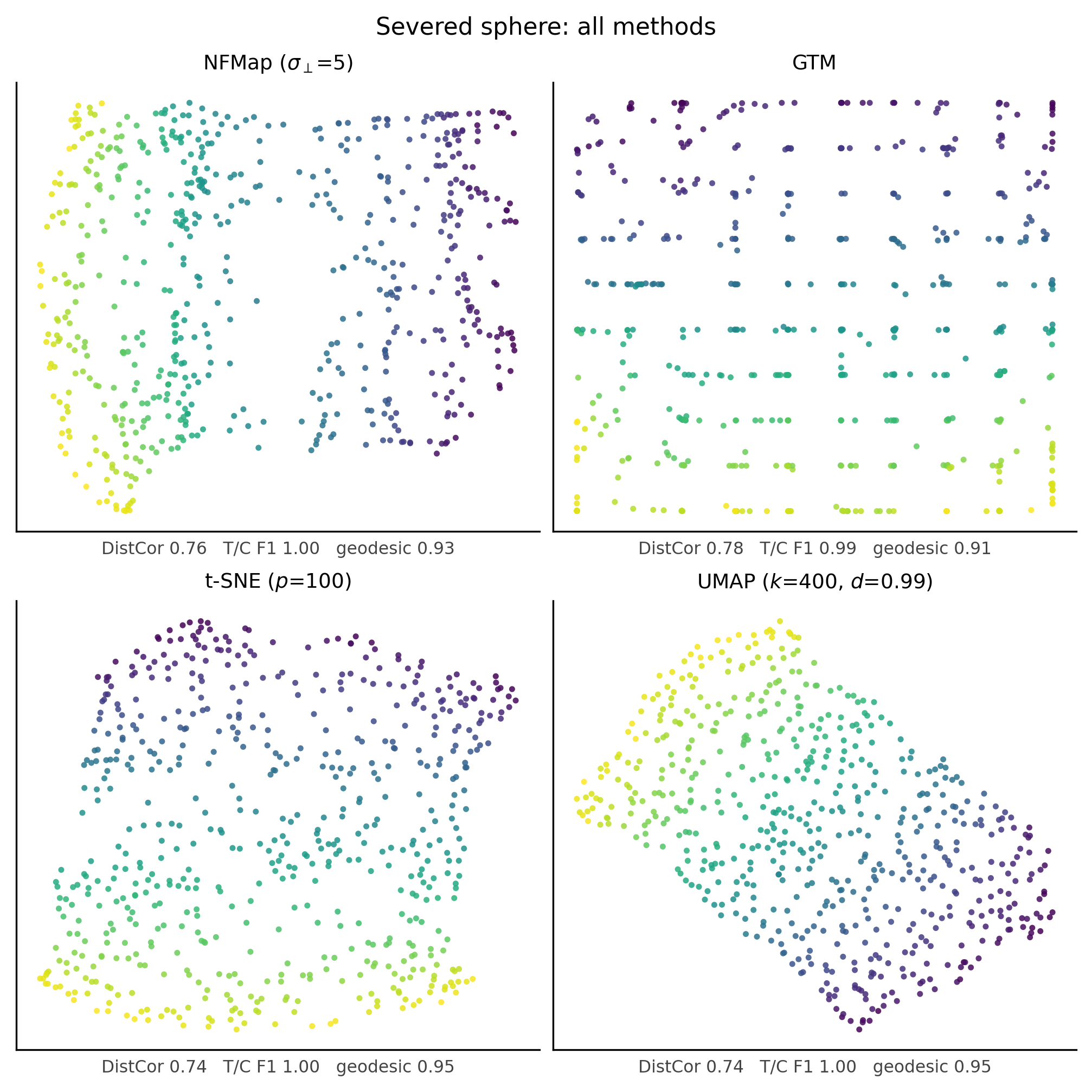}
  \caption{\textbf{Severed sphere manifold embedding.} \nfmap{} and baselines at their Table~\ref{tab:comparison} settings.}
  \label{fig:maps_severedsphere}
\end{figure}

\begin{figure}[H]
  \centering
  \includegraphics[width=\linewidth]{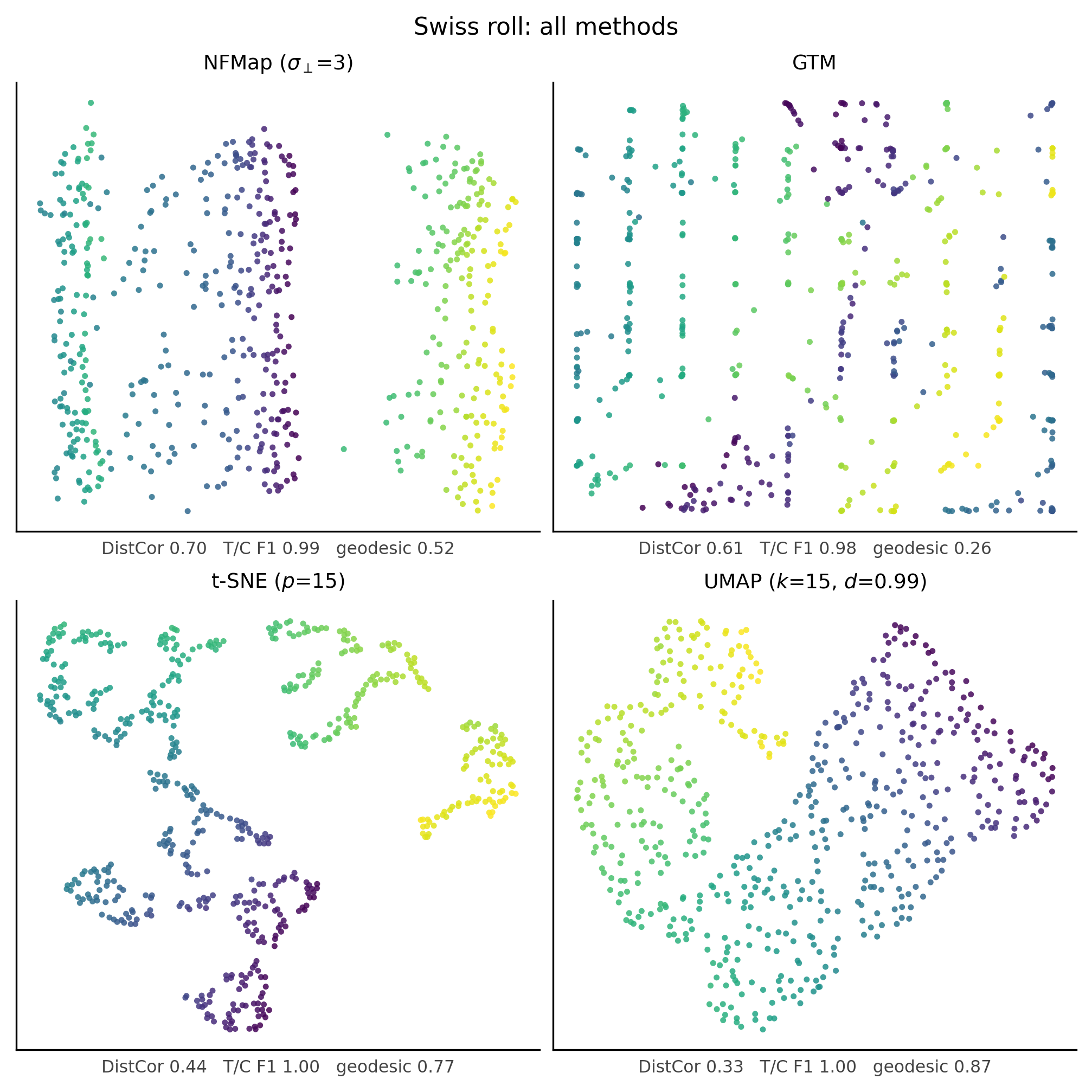}
  \caption{\textbf{Swiss roll manifold embedding.} \nfmap{} and baselines at their Table~\ref{tab:comparison} settings.}
  \label{fig:maps_swissroll}
\end{figure}

In this section, we provide full qualitative map comparisons for the remaining synthetic 3-D benchmark manifolds described in Section~\ref{sec:benchmarks}: the Severed Sphere (Figure~\ref{fig:maps_severedsphere}) and the Swiss Roll (Figure~\ref{fig:maps_swissroll}). 

All baselines (GTM, t-SNE, and UMAP) are depicted at their highest geodesic distance correlation configurations, matching the evaluation protocol reported in Table~\ref{tab:comparison}. 

\newpage 

\newpage

\section{Experimental Settings and Hyperparameters}
\label{app:settings}

Table~\ref{tab:settings} details the hyperparameter sweep ranges and selected Pareto-optimal configurations for each baseline and dataset family reported in Table~\ref{tab:comparison}.

\begin{table}[H]
\centering
\footnotesize
\setlength{\tabcolsep}{3.5pt}
\caption{Hyperparameters selected by the sweep for each entry of Table~\ref{tab:comparison} (the reported Pareto point; not fixed in advance). \nfmap{}: $\sigma_\perp$ / grid\_bounds, swept over 14 $\sigma_\perp$ values from $10^{-3}$ to $10^{3}$ and grid\_bounds 12 and 24 (number of grid nodes fixed at 100). t-SNE: perplexity. UMAP: n\_neighbors / min\_dist. GTM: not swept, one configuration ($10\times10$ nodes, $6\times6$ RBF centres, RBF width factor 0.3).}
\label{tab:settings}
\begin{tabular}{lccc}
\toprule
 & \nfmap{} & t-SNE & UMAP \\
\midrule
\multicolumn{4}{l}{\textit{Chemistry}} \\
\quad BACE & 0.3 / 24 & 200 & 5 / 0.99 \\
\quad BBBP & 3 / 24 & 200 & 5 / 0 \\
\quad ESOL & 0.01 / 24 & 300 & 5 / 0.1 \\
\quad DRD2 & 0.3 / 24 & 300 & 5 / 0.5 \\
\quad EGFR & 0.3 / 24 & 300 & 5 / 0 \\
\multicolumn{4}{l}{\textit{3-D manifolds}} \\
\quad Swiss roll & 3 / 24 & 15 & 15 / 0.99 \\
\quad S-curve & 100 / 12 & 50 & 100 / 0.5 \\
\quad Severed sphere & 5 / 12 & 100 & 400 / 0.99 \\
\multicolumn{4}{l}{\textit{Other}} \\
\quad Digits & 10 / 12 & 400 & 400 / 0.5 \\
\quad Iris & 1 / 12 & 100 & 15 / 0.1 \\
\quad Cancer & 5 / 12 & 200 & 15 / 0.1 \\
\quad Wine & 1 / 12 & 100 & 100 / 0.1 \\
\bottomrule
\end{tabular}
\end{table}

\end{document}